\documentclass[10pt,twocolumn,letterpaper]{article}

\usepackage{iccv}
\usepackage{times}
\usepackage{epsfig}
\usepackage{graphicx}
\usepackage{amsmath}
\usepackage{amssymb}
\usepackage{tabularx,booktabs}
\usepackage{multirow}
\usepackage{arydshln}
\usepackage{flushend}

\usepackage[breaklinks=true,bookmarks=false]{hyperref}

\iccvfinalcopy 

\def\iccvPaperID{****} 

\ificcvfinal\fi

\begin{document}

\title{Resolution based Feature Distillation for Cross Resolution Person Re-Identification}

\author{Asad Munir\\
University of Udine\\
Udine, Italy\\
{\tt\small asad.munir@uniud.it}
\and
Chengjin Lyu\\
TELIN-IPI, Ghent University - imec\\
Ghent, Belgium\\
{\tt\small chengjin.lyu@ugent.be}

\and
Bart Goossens\\
TELIN-IPI, Ghent University - imec\\
Ghent, Belgium\\
{\tt\small bart.goossens@ugent.be}

\and
Wilfried Philips\\
TELIN-IPI, Ghent University - imec\\
Ghent, Belgium\\
{\tt\small wilfried.philips@ugent.be}

\and
Christian Micheloni\\
University of Udine\\
Udine, Italy\\
{\tt\small christian.micheloni@uniud.it}
}

\maketitle
\ificcvfinal\thispagestyle{empty}\fi

\begin{abstract}
Person re-identification (re-id) aims to retrieve images of same identities across different camera views. Resolution mismatch occurs due to varying distances between person of interest and cameras, this significantly degrades the performance of re-id in real world scenarios. Most of the existing approaches resolve the re-id task as low resolution problem in which a low resolution query image is searched in a high resolution images gallery. Several approaches apply image super resolution techniques to produce high resolution images but ignore the multiple resolutions of gallery images which is a better realistic scenario. In this paper, we introduce channel correlations to improve the learning of features from the degraded data. In addition, to overcome the problem of multiple resolutions we propose a  Resolution based Feature Distillation (RFD) approach. Such an approach learns resolution invariant features by filtering the resolution related features from the final feature vectors that are used to compute the distance matrix. We tested the proposed approach on two synthetically created datasets and on one original multi resolution dataset with real degradation. Our approach improves the performance when multiple resolutions occur in the gallery and have comparable results in case of single resolution (low resolution re-id). 
\end{abstract}

\begin{figure}[h!]
\begin{center}
\includegraphics[width=0.45\textwidth]{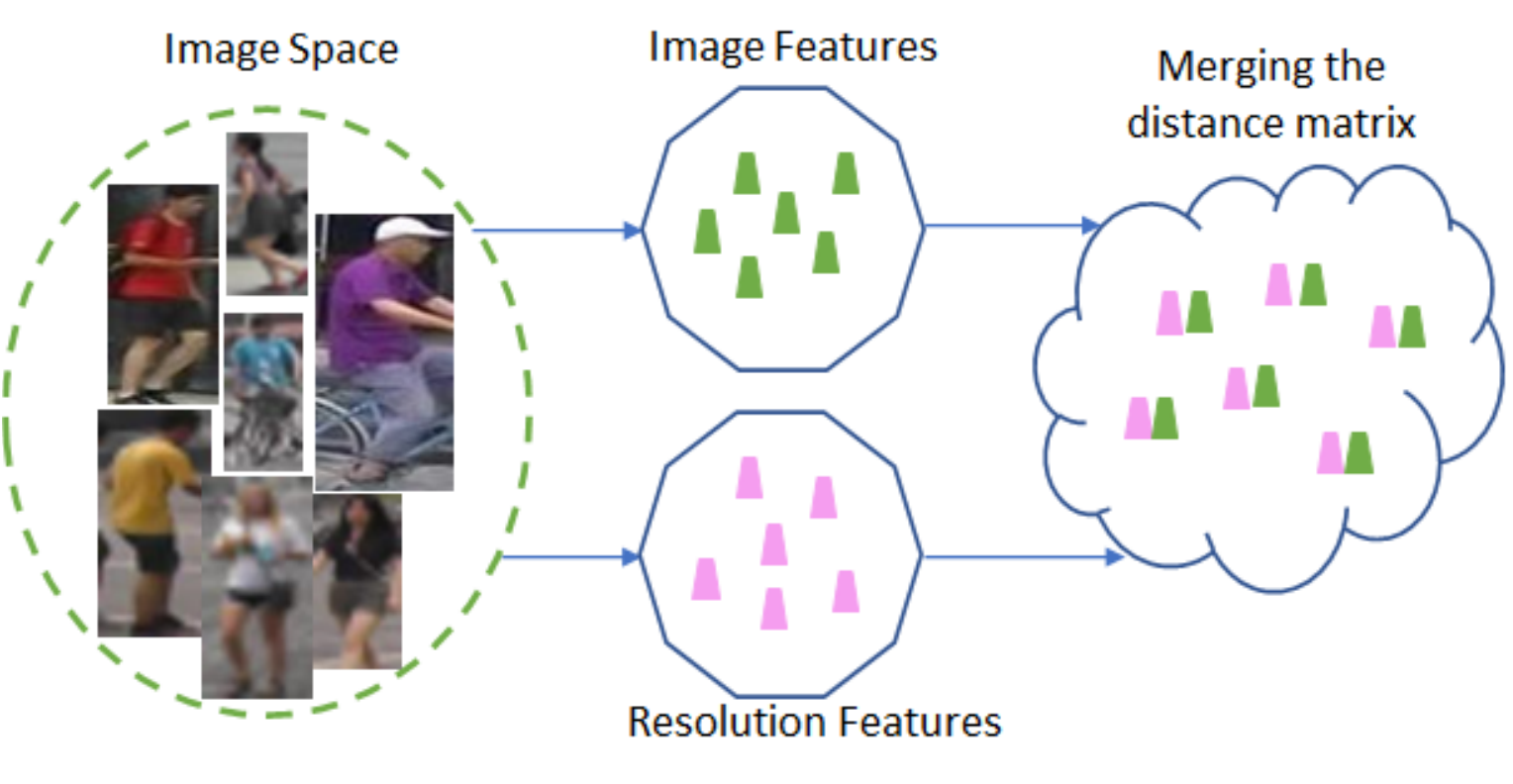}
\end{center}
   \caption{From the input degraded data image and resolution based features are learned by the proposed baseline and their distance matrices are merged to get the final distance matrix.}
\label{fig1}
\label{fig1}
\end{figure}

\section{Introduction}
The task of Person re-identification (re-id) aims to match the images of same person across multiple cameras or views and is an active research topic in the field of computer vision having a long range of applications like video surveillance, person tracking, person searching and computational forensics \cite{micheloni2010intelligent,chen}. The presence of illumination changes, occlusions, background clutter and viewpoint changes makes re-id a challenging task for practical applications.

With a rapid advancement and success of deep learning and convolutional neural networks (CNNs), many learning based approaches \cite{myicip,mgn,pyrnet,sft,abd,IDCL,myicpr} have been proposed for re-id. These methods achieve promising results but assume that query and gallery images have similar resolutions (e.g. high resolution). However, this assumption is not true for realistic scenarios since image resolutions would vary drastically because query images captured by the surveillance cameras are often of low resolution (LR). Usually, the gallery images are chosen beforehand and are of high resolution (HR). This creates a non-trivial resolution mismatching problem due to the direct matching of LR query images with HR gallery images. 

To address low resolution person re-id, most existing methods develop image super resolution (SR) based solutions \cite{sr1,csr,intact,pri,caip19} by converting the LR images into HR images and perform re-id. However, the limitation of such methods is that they operate on known upscaling factors to generate the HR images. A specific SR model is required for each scaling factor which limits generalization (e.g. both source and target resolutions must be known). Several generative adversarial network (GAN) based methods \cite{cr,rain,csr} are also proposed to resolve the above limitations. These methods learn all the degradation into a single network by using adversarial training mechanism and make the learned features independent of the resolution. Although these methods produce promising results, they are computationally expensive and unable to capture highly degraded images which limits exploitation in re-id. Some image super resolution works \cite{srnet,second} used attention mechanisms to improve the learning ability of the network from degraded data and hence provide noise free sharp features to generate a super resolved image. Benefiting from such mechanism, we compute multiple channels correlation in the baseline network \cite{mgn} to learn enhanced features in the presence of degraded data.  

Along with the above mentioned limitations, such methods also work on the assumption that all the gallery images are HR images. In a better realistic scenario, the gallery sets are also collected in the form of LR images (e.g. persons at different distances from the camera) so there should be multiple resolutions even in the gallery images instead of a single resolution (HR). 

In this paper, we propose a Resolution based Feature Distillation (RFD) approach to compute resolution invariant features to solve the multi resolution gallery set problem. First we train a feature baseline (B-F) based on MGN \cite{mgn} approach with several modifications and train a resolution baseline (B-R) to distinguish each resolution present in the synthetic dataset. For the real multi resolution dataset we propose pseudo labeling of the resolutions then train the B-R. In the end we filter out the resolution variant features by combining the distance matrix from both the baselines. The overview of the proposed approach is shown in Fig. \ref{fig1}. The paper has the following contributions:

\begin{itemize}
    \item Improve the baseline network with the addition of multiple channels correlation modules to learn better representations from degraded data.
    \item Propose a resolution based feature distillation with feature and resolution baselines to match the features of different resolution to perform the cross resolution re-id.
    \item Adopt a new and more realistic scenario assuming that the gallery images are also collected in LR form. Thus, the LR query image is matched with multiple resolutions (HR and LR) instead of a single resolution (HR).
\end{itemize}
With the above mentioned contributions, we performed the experiments on two synthetically created re-id datasets and one real dataset with multiple resolutions. Our feature baseline produces competitive results on the low resolution re-id (single resolution in gallery set) with the other state of the art methods. The proposed RFD approach improves the results in the case of multiple resolutions in the gallery set when compared to single baseline (B-F).

\section{Related Work}
\subsection{Person re-id}
A variety of person re-id methods have been proposed to address various challenges in re-id task. Several approaches \cite{sft,mgn} are based on resnet-50 structure along with modifications suitable for person-re-id. Many methods \cite{IDCL} have been developed based on different loss functions to learn the most discriminative features for re-id. Another \cite{myicip,joint,crossdom,style} trend focuses on learning pose and domain variations by using generative adversarial networks (GANs). Several efforts \cite{abd, myicpr} have been made to learn long range dependencies in the features by introducing attention mechanism in the networks. Recently, many works introduce unsupervised person re-identification \cite{unsupervised1,unsupervised2}.However, these methods typically assume that the query and gallery images have similar resolution (HR) which is not practically true in real world applications.

\subsection{Cross resolution person re-id}
Several methods have been proposed to solve the resolution mismatch problem in person re-id. Li et al. \cite{li} carry out cross scale image domain alignment and multi scale distance metric learning jointly. Jing et al. \cite{jing} perform a mapping between HR and LR images with the help of a semi-coupled low-rank dictionary. Wang et al. \cite{wang} develop a framework to learn a discriminative scale-distance function by varying the image scale of LR images when matching with HR images. All these approaches adopt handcrafted descriptors which cannot enhance the person re-id performance as compared to CNNs.\\
Recently, several CNN-based models have been proposed to perform cross resolution person re-id. SING \cite{sing} proposed a network composed of several SR sub-networks and a person re-id module for low resolution person re-id. RIPR \cite{ripr} jointly trained foreground focus SR module and resolution invariant feature extractor to learn high resolution features for LR person re-id. A number of GAN based methods have been introduced and they perform significantly better than the CNN-based methods. CSR-GAN \cite{csr} cascades multiple GANs to progressively recover the LR image details. In all the above methods, SR models are a necessary part of the training and some of them use pretrained SR models. RAIN \cite{rain} and CRGAN \cite{cr} align the feature distributions of LR and HR images to improve the LR person re-id performance. MSA \cite{msa} performs SR, denoising and re-id separately to achieve the final performance. \cite{pri,intact} use image super resolution techniques and predict the scaling factors to make the query image high resolution when matching with high resolution gallery images.\\
All these methods adopt single resolution (HR) gallery set while the proposed approach considers multiple resolutions gallery sets. Our approach uses a resnet-50 \cite{resnet50} based MGN \cite{mgn} like baseline for extracting the features and then filters out the resolution dependent features to create a strong descriptor to match LR and HR images.

\section{Proposed Method}
In this section, we provide the approach overview, network architecture and resolution based feature distillation (RFD) technique.

\subsection{Notations and Overview}
Let $X_H=\big\{x_i^H\big\}_{i=1}^{N}$ be a set of $N$ HR training images, with corresponding identities labels $Y_H\big\{y_i^H\big\}_{i=1}^{N}$,  acquired by a camera network where $x_i^H \in \mathbb{R}^{H \times W \times 3}$ and $y_i^H \in \mathbb{R}$ are the $i^{th}$ HR image and its label respectively. To enable the model to learn different resolutions, we generate the synthetic LR image set $X_L=\big\{x_i^L\big\}_{i=1}^{N}$ by downsampling and upsampling (bilinear) back to the original image size each image in $X_H$ i.e. $x_i^L \in \mathbb{R}^{H \times W \times 3}$. The $Y_L$ labels for the corresponding $X_L$ images are the same of $Y_H$. \\
Low resolution person re-id requires to retrieve images of the same identity from a HR gallery images of different cameras by giving a LR probe image. While in this work we make a cross resolution person re-id by building a gallery set of multiple resolutions instead of single resolution (HR). During the training stage, a feature baseline B-F is used to extract the features from the images of all resolutions and persons identities are predicted by exploiting classification and metric learning losses. The networks details and the loss functions are explained in the next section. A resolution baseline B-R is trained to classify the resolutions present in the dataset. In the testing stage, We compute the distance matrix from each baseline and then merge them to get the final distance matrix. This final distance matrix is used to perform cross resolution person re-id.

\begin{figure*}
\begin{center}
\includegraphics[width=0.98\textwidth]{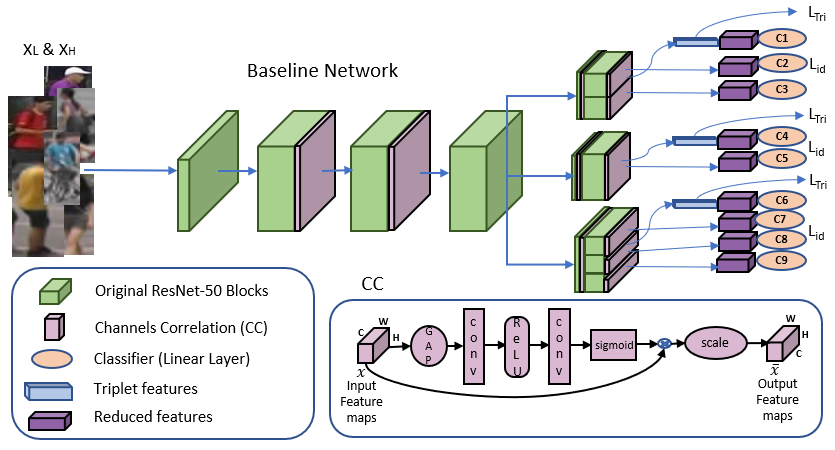}
\end{center}
   \caption{Architecture of the baseline and overview of the proposed Resolution based Feature Distillation (RFD) approach.}
\label{fig2}
\end{figure*}

\subsection{Network Architecture}
Recent works have shown that Convolutional Neural Networks (CNNs) are efficient for learning deeper and robust feature representations from images and are accurate to train if they have shorter connection between layers. Relying on such outcomes, we define ResNet-50 \cite{resnet50} as our baseline network with several adjustments. We follow the MGN \cite{mgn} architecture and introduce the final convolutional block  three times to make global and local modules. Concerning the global module, we modify the stride (stride=1) of last downsampling block to make the spatial size of the convolutional features larger before global average pooling. Channel correlations are computed at each stage of the network as in \cite{srnet} to make the learned feature sharp and discriminative. Unlike \cite{srnet}, we introduce the channel attention modules after each stage of the network instead of using them only before residual connection as shown in Fig. \ref{fig2}.
Cross entropy loss is used to predict the identity of each person and all the cross entropy losses from each branch are added to get the final loss which is given by:
\begin{equation}
   L_{id} = -\sum_{b=1}^{B_i}\sum_{c=1}^{C}\log(p_b(c))q_b(c)
\label{eq1}
\end{equation}
where $B_i$ is the number of classification branches.
Hard mining triplet loss from each branch is added to achieve the final loss to make the features discriminative and it is computed as:
\begin{equation}
   L_{tri} = \sum_{b=1}^{B_t}max(\left\Vert x_b^a - x_b^p\right\Vert - \left\Vert x_b^a - x_b^n\right\Vert + \alpha,0)
\label{eq2}
\end{equation}
where $B_t$ is the number of branches to compute triplet loss. This loss ensures that the representation of the positive sample is closer, by at least a margin $\alpha$, to the anchor sample than to the negative
one. For both the baselines, we use the same structure and modifications discussed above. Now we review channels correlation which are computed with the help of channel attention mechanism.

\subsection{Channels Correlation}
Since person re-identification is applied to surveillance cameras, commonly real scenarios and used datasets consist of blurry and noisy images. Most of the existing methods are unable to grasp deep salient features from them. To build a stronger descriptor against such a degradation, noise free and distinct feature learning is required. To fulfill this objective, we introduce several channel attention modules to compute channels correlation consistently during the feature learning process. 

Let $K=[k_1, k_2,...,k_C]$ be the learned set of filter kernels for $C$ output channels with $k_l$ being the parameters of the $l^{th}$ filter in a general convolution operation. The output from this convolution operation can be written as $U=[u_1, u_2,..., u_C]$, where
\begin{equation}
   u_l = k_l * X = \sum_{n=1}^{C'}k_l^n * x^n
\label{eq3}
\end{equation}
In the above equation, $k_l=[k_l^1, k_l^2,..., k_l^{C'}]$, $X=[x^1, x^2,..., x^{C'}]$ ($X$ being the input feature maps and $C'$ is the number of input channels). The convolution operation is denoted by * and 2D spatial kernel $k_c^n$ represents a single channel of $k_l$ which interacts with the corresponding channel of $X$. The output of the convolutional layers is obtained through a channel-wise sum of the computed feature values. Therefore, the channel dependencies are introduced along with the spatial correlation captured by the convolutional filters in the learned weights. We follow the work in \cite{srnet,sen} for computing these channel dependencies (correlations) but apply them at compacted features (convolutional block) instead at residual connections (used in \cite{srnet,sen}). \\
Each unit of the output $U$ is unable to exploit contextual information outside of its region because the convolution operation has a local receptive field. To resolve this issue, global spatial information is squeezed into a channel descriptor. This operation generates channel-wise statistics and is achieved by using global average pooling. A statistic $z \in \mathbb{R}^C$ can be generated by shrinking $U$ through the spatial dimension $H \times W$. The $l^{th}$ element of $z$, computed by global average pooling, can be written as:
\begin{equation}
   z_l = \frac{1}{H\times W}\sum_{i=1}^{H}\sum_{j=1}^{W}u_l(i,j)
\label{eq4}
\end{equation}
For better modeling channel-wise dependencies, the learned function must have the ability to capture the nonlinear interaction between channels and permit multiple channels to oppose one-hot activation. The sigmoid activation fulfills these requirements and can be written as:
\begin{equation}
   n = \sigma(g(z,W)) = \sigma(W_2\delta(W_1z))
\label{eq5}
\end{equation}
where $W_1\in \mathbb{R}^{\frac{C}{r}\times C}$ are the parameters of the dimensionality reduction layer and  $W_2\in \mathbb{R}^{C\times \frac{C}{r}}$ are the parameters of dimensionality-increasing layer while $\delta$ denotes the ReLU function and $r$ is the reduction ratio. Two $1\times 1$ convolution layers implement $W_1$ and $W_2$ around the non-linearity. The final output of the channel attention is obtained by rescaling the output $U$ by means of the activations:
\begin{equation}
   \bar x_l = n_l \cdot u_l 
\label{eq6}
\end{equation}
where $\bar X=[\bar x_1, \bar x_2,..., \bar x_C]$. The dot product refers to channel-wise multiplication of feature maps $u_l\in \mathbb{R}^{H\times W}$ and the scalar $n_l$. The overall operation of the channel attention for computing channels correlation is shown in Fig. \ref{fig2} and it helps to boost feature discrimination.

\subsection{Resolution based Feature Distillation}
In this section, first we perform the learning of resolutions similarity as a resolution re-id problem. This task is performed by using the baseline B-F to compute the resolution features and provide a classification between the available resolutions. However, instead of outputting a regression result or a classified direction, resolution re-id outputs an embedding that can be used to compute resolutions similarity. To train the baseline B-R we need the resolution labels which are explained in next section. In testing stage, for an $x$ image we retain the resolution features $f_r(x)$ from the baseline B-R. Then calculate the distance between the two images $x_i$ and $x_j$ is computed as:
\begin{equation}
   D_r(x_i, x_j) = \frac{f_r(x_i)\cdot f_r(x_j)}{\left\Vert f_r(x_i)\right\Vert \left\Vert f_r(x_j)\right\Vert}
\label{eq7}
\end{equation}
The distance matrix from the baseline B-F is computed as the euclidean distance between query and gallery images and then merged with the distance in eq.\ref{eq7} to compute the final distance matrix for person re-id.
The final distance matrix provides resolution invariant matching of features and it is given by:
\begin{equation}
   D(x_i, x_j) = D_f(x_i, x_j) - \lambda D_r(x_i, x_j)
\label{eq8}
\end{equation}
where $D_f$ is the distance matrix calculated from the baseline B-F similar to eq.\ref{eq7} and $\lambda$ is the scaling parameter (we use $\lambda=0.1$ for all our experiments). This distillation process enhance the performance when we have multiple resolutions in the gallery set as well as in the query set and is shown in Fig. \ref{fig3}.

\begin{figure}[h]
\begin{center}
\includegraphics[width=0.45\textwidth]{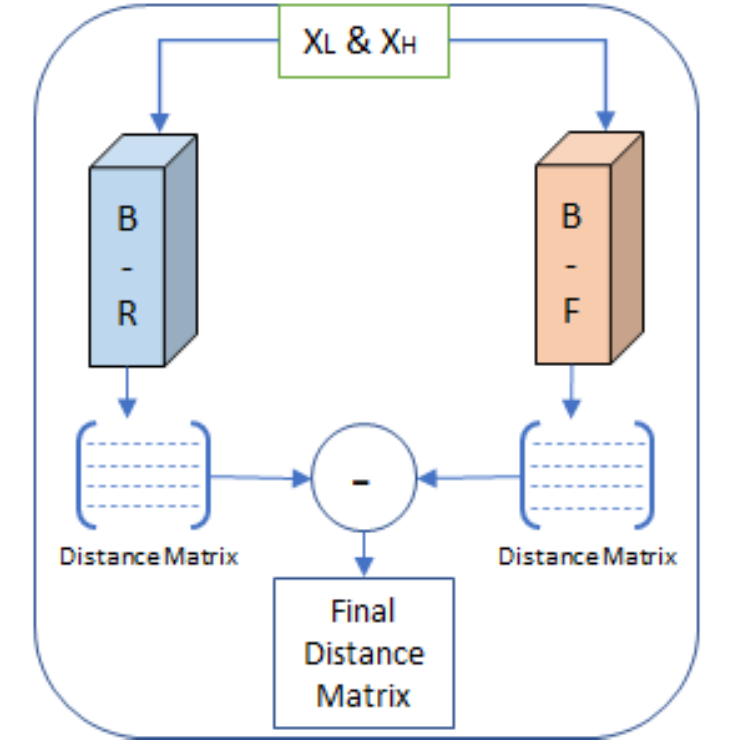}
\end{center}
   \caption{Proposed Resolution based Feature Distillation (RFD) training mechanism. $x_L$ and $x_H$ are the multiple low resolution and high resolution images. B-F and B-R are the baselines trained with person ids and resolution ids respectively.}
\label{fig3}
\label{fig3}
\end{figure}

\subsection{Resolution pseudo labeling}
The training of the baseline B-R for multiple resolutions requires labels for each resolution in the dataset. For synthetic datasets we have known downsampling scaling factors and we used them as training labels for baseline B-R. For synthetic datasets we have $4$ different classes (labels) which include $3$ downscaling factors i.e.${2,3,4}$ and one original HR resolution. This type of downsampling is not available for the real dataset having multiple different resolutions. So, for them, we propose a pseudo labeling process that computes the total number of pixels for each image. Then we divide these number of pixels into $5$ patches of equal length. Each image within the same patch is represented by single resolution pseudo label.  With this technique, we build the training set with $5$ resolutions for the real dataset.

\section{Experiments}
\subsection{Implementation Details}
We implemented the proposed network using Pytorch. The baselines are build upon a ResNet-50 network pretrained on ImageNet. We modified the network with all the adjustments mentioned in section $3.2$. We optimized the network by using Adam optimizer with momentum $0.9$. We trained the models for $500 \times (\textrm{\textit{number of identities}})$ iteration with the initial learning rate is $2e-4$ and is divided by $10$ after half iterations. We used a batch size of 32 (having  $8$ identities with $4$ images each) and $(\textrm{\textit{number of resolutions}}) \times 16$ for baseline B-F and baseline B-R respectively. All the images are resized to $384\times 128$ with random horizontal flipping and random erasing data augmentations. The dropout probability is set to $0.5$ and the weight decay is $5e-4$.
\subsection{Datasets}
we evaluated the proposed method on three datasets which are described as follows:\\
\textbf{MLR-market} dataset consists of $32668$ images of $1501$ identities captured from $6$ different cameras. Training and testing set consists of $751$ and $750$ identities respectively. We follow SING \cite{sing} to create the MLR-market1501 dataset.\\
\textbf{MLR-DukeMTMC-reID} dataset contains $36411$ images of $1404$ identities with $8$ camera views. Following SING \cite{sing} and $702/702$ splits for training and testing, we create MLR-DukeMTMC-reID dataset.\\
\textbf{CAVIAR} is a real multi resolution dataset and is composed of $1220$ images of $72$ persons taken from $2$ different cameras. We discard $22$ identities those appear in one camera and split the dataset into two non-overlapping halves by following SING \cite{sing}.

\subsection{Comparison with state-of-the-art}
Table \ref{table1}, \ref{table2} and \ref{table3} show the results of the proposed method and its comparison with other state of the art methods on two synthetic datasets and one real dataset. We refer MLR-Market and MLR-Duke as synthetic datasets because the degraded process is known in datasets while the CAVIAR dataset has unknown degradation. We computed rank-1 (R1), rank-5 (R5) and rank-10 (R10) accuracy by using our baseline B-F. We used bicubic downsampling to generate the low resolution images with a randomly selected scaling factor of ${2,3,4}$. The results in Table \ref{table1}, \ref{table2} and \ref{table3} are computed for low resolution person re-id in which the query image is randomly down sampled and the gallery set has all the HR images. Like for the existing methods, we make random splits of the data for $10$ times and then take the average to calculate the final scores. The performance of the proposed method is significant when compared to the other state of the art methods. 

\begin{table}[h]
\centering
\resizebox{\columnwidth}{!}{\begin{tabular}{|c|c|ccc|}
    \hline
    \multirow{2}{*}{\textbf{Methods}} &
    \multirow{2}{*}{\textbf{Reference}} &
    \multicolumn{3}{c|}{\textbf{MLR-Market}} \\
    \cline{3-5}
    & & \textbf{\textit{R1}} & \textbf{\textit{R5}} & \textbf{\textit{R10}} \\
    \hline
    \hline
    SING \cite{sing} & AAAI18 &  74.4 & 87.8 & 91.6  \\
    CSR-GAN \cite{csr} & IJCAI18 & 76.4 & 88.5 & 91.9  \\
    CamStyle \cite{cam} & CVPR18 & 74.5 & 88.6 & 93.0  \\
    FD-GAN \cite{fd} & NeurIPS18 & 79.6 & 91.6 & 93.5   \\
    RIPR \cite{ripr} & IJCAI19 & 66.9 & 84.7 & -  \\
    CRGAN \cite{cr} & ICCV19 & 83.7 & 92.7 & 95.8  \\
    MSA \cite{msa} & IEEE Access20 & 68.3 & 85.7 & -  \\
    INTACT \cite{intact} & CVPR20 & \textcolor{red}{88.1} & \textcolor{blue}{95.0} & \textcolor{blue}{96.9}  \\
    PRI \cite{pri} & ECCV20 & \textcolor{red}{88.1} & 94.2 & 96.5  \\
    \hdashline
    baseline  & - & 84.1 & 92.3 & 95.9  \\
    baseline B-F  & - & 85.5 & 94.1 & 96.0  \\
    (B-F+RFD) Proposed  & - & \textcolor{blue}{86.9} & \textcolor{red}{95.6} & \textcolor{red}{97.4}  \\
    \hline
\end{tabular}}
\caption{Comparisons of the proposed method with the state-of-the-art re-id methods on MLR-Market dataset. The best and second best results are highlighted in \textcolor{red}{red} and \textcolor{blue}{blue} respectively.}
\label{table1}
\end{table}

Table \ref{table1} shows the results and comparison of the proposed method with state of the art methods of MLR-Market dataset. The proposed contributions improved the R1 accuracy by $2.8\%$ when compared to the baseline. The R1 accuracy of the proposed method is second highest in the table when compared to the state of the art methods but the improvement in the baseline is quit significant. The results for MLR-Duke dataset is presented in Table \ref{table2} and the performance of the proposed approach is superior to the state of the art methods in terms of R1, R5 and R10 accuracy. The proposed approach performs better and uses training data only which contains low and high resolution images and no extra data generated or any super resolution models are introduced. 

\begin{table}[h]
\centering
\resizebox{\columnwidth}{!}{\begin{tabular}{|c|c|ccc|}
    \hline
    \multirow{2}{*}{\textbf{Methods}} &
    \multirow{2}{*}{\textbf{Reference}} &
    \multicolumn{3}{c|}{\textbf{MLR-Duke}} \\
    \cline{3-5}
    & &\textbf{\textit{R1}} & \textbf{\textit{R5}} & \textbf{\textit{R10}} \\
    \hline
    SING \cite{sing} & AAAI18 &  65.2 & 80.1 & 84.8  \\
    CSR-GAN \cite{csr} & IJCAI18 & 67.6 & 81.4 & 85.1  \\
    CamStyle \cite{cam} & CVPR18  & 64.0 & 78.1 & 84.4  \\
    FD-GAN \cite{fd} & NeurIPS18 &  67.5 & 82.0 & 85.3  \\
    CRGAN \cite{cr} & ICCV19 & 75.6 & 86.7 & 89.6  \\
    MSA \cite{msa} & IEEE Access20  & 79.1 & 90.0 & -  \\
    INTACT \cite{intact} & CVPR20  & 81.2 & 90.1 & \textcolor{blue}{92.8}  \\
    PRI \cite{pri} & ECCV20  & \textcolor{blue}{82.1} & \textcolor{blue}{91.1} & \textcolor{blue}{92.8}  \\
    \hdashline
    baseline  & - & 81.2 & 90.1 & 91.9  \\
    baseline B-F  & - & 82.0 & 90.8 & 92.7  \\
    (B-F+RFD) Proposed  & -  & \textcolor{red}{82.9} & \textcolor{red}{92.0} & \textcolor{red}{94.0}  \\
    \hline
\end{tabular}}
\caption{Results and comparisons of the proposed method with the state-of-the-art re-id methods on MLR-Duke dataset. The best and second best results are highlighted in \textcolor{red}{red} and \textcolor{blue}{blue} respectively.}
\label{table2}
\end{table}

The results and comparison of the proposed approach on real world degraded dataset CAVIAR is shown in Table \ref{table3}. We obtained highest results among all other methods in the presence of real world degradation. The proposed RFD enhanced the R1 accuracy by $3.3\%$ when compared with baseline B-F and the reason for such an improvement is the proposed pseudo labeling technique which separate resolutions through baseline B-R. The CAVIAR dataset is a very small dataset as compared to MLR-Market and MLR-Duke so improving the rank 1's of few queries have a greater impact on the performance.

\begin{table}[h]
\centering
\resizebox{\columnwidth}{!}{\begin{tabular}{|c|c|ccc|}
    \hline
    \multirow{2}{*}{\textbf{Methods}} &
    \multirow{2}{*}{\textbf{Reference}} &\multicolumn{3}{c|}{\textbf{CAVIAR}}\\
    \cline{3-5}
    & &\textbf{\textit{R1}} & \textbf{\textit{R5}} & \textbf{\textit{R10}} \\
    \hline
    \hline
    SING \cite{sing} & AAAI18 & 33.5 & 72.7 & 89.0   \\
    CSR-GAN \cite{csr} & IJCAI18 & 34.7 & 72.5 & 87.4  \\
    CamStyle \cite{cam} & CVPR18 & 32.1 & 72.3 & 85.9   \\
    FD-GAN \cite{fd} & NeurIPS18 & 33.5 & 71.4 & 86.5  \\
    RIPR \cite{ripr} & IJCAI19 & 36.4 & 72.0 & -  \\
    CRGAN \cite{cr} & ICCV19 & 42.8 & 76.2 & 91.5  \\
    INTACT \cite{intact} & CVPR20 & 44.0 & 81.8 & 93.9  \\
    PRI \cite{pri} & ECCV20 & \textcolor{blue}{45.2} & 84.1 & \textcolor{blue}{94.6}  \\
    \hdashline
    baseline  & - & 42.1 & 87.6 & 92.9  \\
    baseline B-F  & - & 44.3 & \textcolor{blue}{88.4} & 94.2  \\
    (B-F+RFD) Proposed  & - & \textcolor{red}{47.6} & \textcolor{red}{89.2} & \textcolor{red}{96.0}  \\
    \hline
\end{tabular}}
\caption{Comparisons of the proposed method with the state-of-the-art re-id methods on the real world multi resolution CAVIAR dataset. The best and second best results are highlighted in \textcolor{red}{red} and \textcolor{blue}{blue} respectively.}
\label{table3}
\end{table}

\begin{figure*}[h]
\centering
\includegraphics[width=\textwidth]{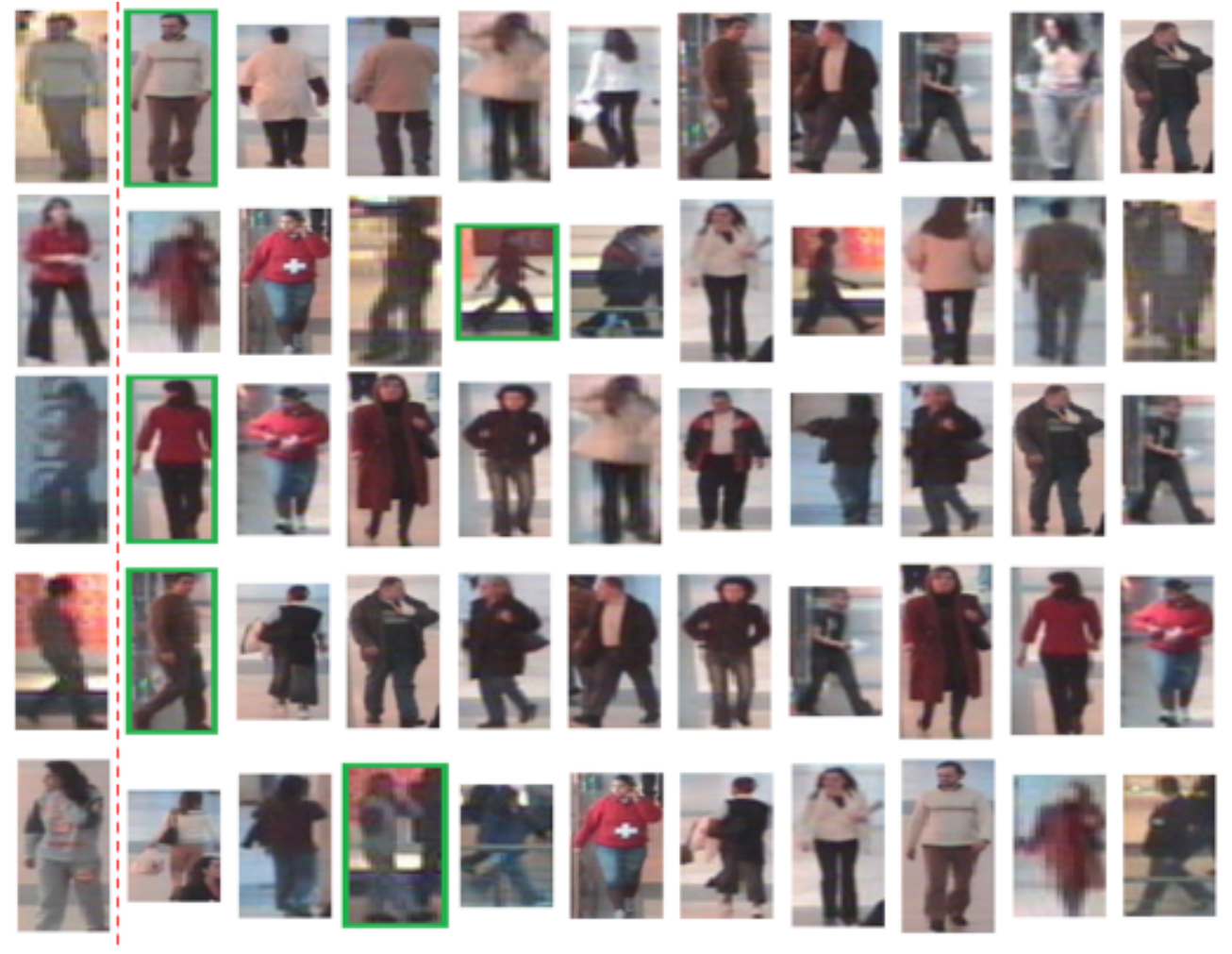}
\caption{Visual results extracted by the proposed method. Query image with its first ten matches in the gallery for CAVIAR dataset.} 
\label{fig4}
\end{figure*}

\subsection{Ablation Study}

We performed an ablation study for the proposed approach in Table \ref{table4} and \ref{table5}. We created the synthetic data (low resolution images) by using two types of interpolation bicubic and bilinear for MLR-Market and MLR-Duke dataset. We performed the experiments and recorded the results in first two rows of Table \ref{table4}. The performance remains almost similar with both interpolations. Third and fourth rows show the results on single and multi resolution gallery set respectively and results are without using RFD method. We generated the multi resolution gallery set by randomly downsample images from synthetic datasets and by randomly pick one image for each identity from all the resolutions for CAVIAR dataset. The performance for CAVIAR dataset is recorded in the first and second rows of the Table \ref{table5} for single and multiple resolutions in gallery set. This multi resolution causes performance reduction for synthetic datasets while performs better in the case of real dataset. The reason for this better performance is the multiple HR query images are also present unlike single resolution. The improvement with the proposed RFD for multiple resolutions in gallery is shown in the last row of Table \ref{table2} and \ref{table5} for synthetic and real data respectively. The pseudo labeling significantly enhance the performance for CAVAIR dataset. Rank-5 scores remains almost similar because of the presence of only single image for each query in the gallery set.

\begin{table}[h]
\centering
\resizebox{\columnwidth}{!}{\begin{tabular}{|c|cc|cc|}
    \hline
    \multirow{2}{*}{\textbf{Components}} & 
    \multicolumn{2}{c|}{\textbf{MLR-Market}} &
    \multicolumn{2}{c|}{\textbf{MLR-Duke}} \\
    \cline{2-5}
    & \textbf{\textit{R1}} & \textbf{\textit{R5}}  & \textbf{\textit{R1}} & \textbf{\textit{R5}}  \\
    \hline
    \hline
    Bicubic   & 83.1 & 96.0  & 82.4 & 92.0  \\
    Bilinear   & 82.9 & 95.9  & 82.2 & 92.1  \\
    Single-Reso   & 85.5 & 94.1  & 82.0 & 90.8  \\
    Multi-Reso   & 84.2 & 93.8  & 81.5 & 90.0  \\
    Multi-Reso+RFD   & 85.6 & 94.4  & 83.0 & 91.8  \\
    \hline
\end{tabular}}
\caption{Ablation study of the proposed method on two synthetic datasets MLR-Market and MLR-Duke.}
\label{table4}
\end{table}

\begin{table}[h]
\centering
\resizebox{\columnwidth}{!}{\begin{tabular}{|c|ccc|}
    \hline
    \multirow{2}{*}{\textbf{Components}} & \multicolumn{3}{c|}{\textbf{CAVIAR}} \\
    \cline{2-4}
    & \textbf{\textit{R1}} & \textbf{\textit{R5}} & \textbf{\textit{R10}}  \\
    \hline
    \hline
    Single-Reso  & 44.3 & 88.4 &  94.2 \\
    Multi-Reso  & 55.5 & 90.6  & 91.5  \\
    multi-Reso+RFD  & 56.4 & 90.6  & 92.2 \\
    \hline
\end{tabular}}
\caption{Results of the proposed method for single and multi resolution gallery.}
\label{table5}
\end{table}

\begin{figure}[h]
\centering
\includegraphics[width=0.45\textwidth]{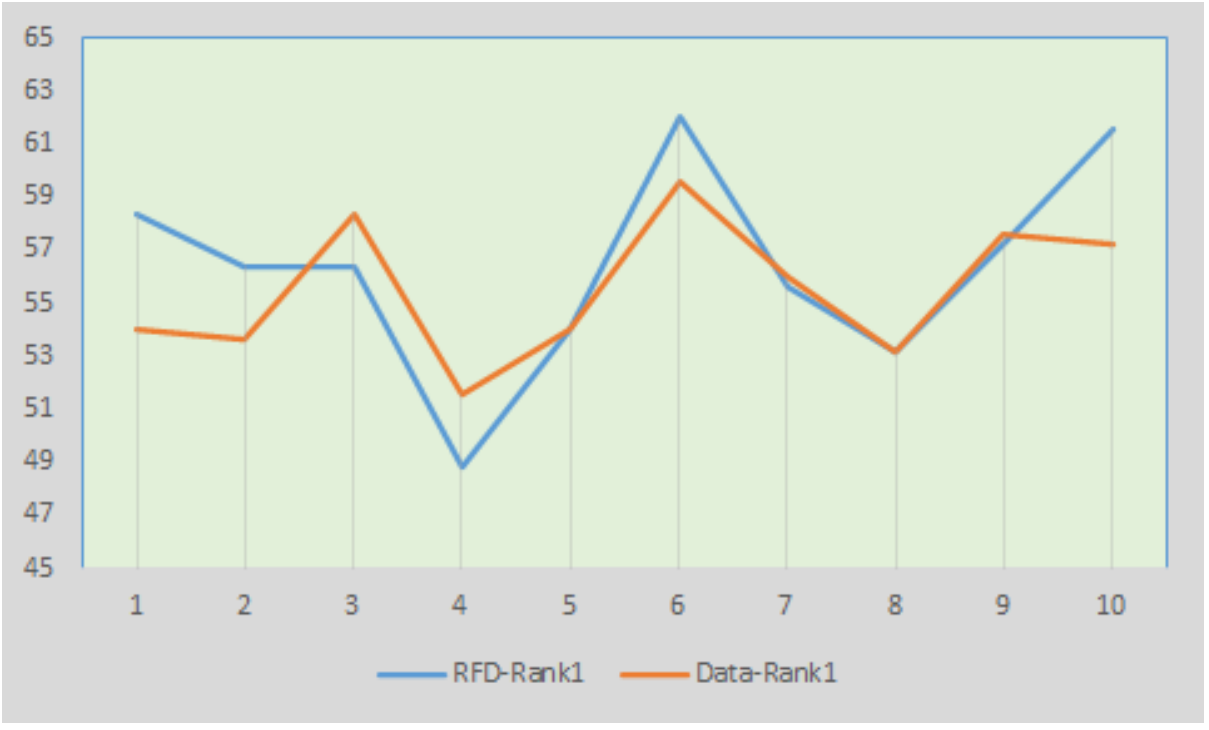}
\caption{The effect of proposed pseudo labeling on CAVIAR dataset. Random splits of the dataset are shown on horizontal axis with their performance on vertical axis.} \label{fig5}
\end{figure}

We presented the effects of the proposed pseudo labeling for CAVIAR dataset in Fig. \ref{fig5}. We generated ten random splits of the dataset for testing and recorded their performance (rank-1 accuracy) along with RFD approach in Fig. \ref{fig5}. Orange and blue lines represent the baseline B-F and RFD performances respectively. Some splits are significantly better while the others have similar performance. We noticed a reduction in the performance of  $3rd$ and $4th$ split which is due to the fact that the gallery and query set in that split do not have much resolution variance. The average scores are higher than the baseline when using RFD for 10 splits of the data.

The visual results of the proposed approach are shown in Fig. \ref{fig4}. Five query images from CAVIAR dataset are presented along with their first ten matches in the gallery set. Query images are on the left side of the red dashed line while right side of the red dashed line are the top 10 rankings from the gallery images. There is only one true match since gallery set only has single images for each query to match. We did not resize the images so the original resolutions have been used and are shown. The green rectangle is used to show the true match in the gallery set. 

\section{Conclusion}
We proposed a resolution based feature distillation (RFD) approach for cross resolution person re-identification. Firstly, we improved a baseline by means of channels correlation to solve the general low resolution person re-id problem (only HR images in the gallery set). We achieved competitive results on three datasets. Secondly, we proposed a resolution based feature distillation technique which filters out the resolution dependent features to compute the final distance matrix for matching. A pseudo labelling technique for computing the resolution label is also introduced to train the RFD. The proposed approach considers the better realistic scenario in which gallery sets contain multiple resolutions instead of single resolution (HR). This is a novel scenario that we think should be further investigated in future studies within the person re-id community. Experimental results have shown how the RFD approach improves with respect to both the baseline and other state of the art approaches.

\section*{Acknowledgement}
This work was supported by EU H2020 MSCA through Project ACHIEVE-ITN (Grant No 765866).

{\small
\bibliographystyle{ieee_fullname}
\bibliography{egbib}
}

\end{document}